%% file: main.tex
\documentclass[letterpaper, 10 pt, conference]{ieeeconf}

\IEEEoverridecommandlockouts                              % This command is only needed if 
\usepackage{graphics} % for pdf, bitmapped graphics files

\usepackage{stmaryrd} % for pdf, bitmapped graphics files
\usepackage{epsfig} % for postscript graphics files
\usepackage{times} % assumes new font selection scheme installed
\usepackage{amsmath} % assumes amsmath package installed
\usepackage{amssymb}  % assumes amsmath package installed
\usepackage{float}
\usepackage{hyperref}
\usepackage{subcaption}
\usepackage{physics}
\usepackage{tikz}
\usepackage{pgfplots}
\usepackage{balance}
\usepackage{tabu}
\usepackage{multirow,booktabs}

\title{\LARGE \bf
Intensity Scan Context: Coding Intensity and Geometry Relations for Loop Closure Detection}

\author{Han Wang, Chen Wang, and Lihua Xie% <-this % stops a space
\thanks{
The work is supported by Delta-NTU Corporate Laboratory for Cyber-Physical Systems under the National Research Foundation Corporate Lab @ University Scheme.
}
\thanks{Han Wang and Lihua Xie are with the School of Electrical and Electronic Engineering,
Nanyang Technological University, 50 Nanyang Avenue, Singapore 639798.
        {\tt\small e-mail: \{wang.han,elhxie\}@ntu.edu.sg}}%
\thanks{Chen Wang is with the Robotics Institute, Carnegie Mellon University, Pittsburgh, PA 15213, USA. {\tt\small e-mail: chenwang@dr.com}}
}

\begin{document}
 
\maketitle
\thispagestyle{empty}
\pagestyle{empty}

%特点：
%这篇文章的主要目的是什么，介绍了一个新的intensity scan context for place recognition, 然后再跟vision做融合以达到一个最好的效果

%global descriptor的问题，code both geometry and intensity information， faster speed， 但是精度不是很高，这就需要视觉来辅佐

%%%%%%%%%%%%%%%%%%%%%%%%%%%%%%%%%%%%%%%%%%%%%%%%%%%%%%%%%%%%%%%%%%%%%%%%%%%%%%%%
\begin{abstract}

\input{body/abstract.tex}

\end{abstract}
%%%%%%%%%%%%%%%%%%%%%%%%%%%%%%%%%%%%%%%%%%%%%%%%%%%%%%%%%%%%%%%%%%%%%%%%%%%%%%%%
\section{INTRODUCTION}
\input{body/Introduction.tex}

\section{RELATED WORK}

\input{body/RelatedWork.tex}

\section{METHODOLOGY}
\input{body/Methodology.tex}

\section{EXPERIMENT EVALUATION}
\input{body/Experiments.tex}

\section{CONCLUSION}
\input{body/Conclusion.tex}

%\section*{ACKNOWLEDGMENT}
%The author would like to thank Dr. Qiu Zhirong for many great suggestions during the course of this research work.
\balance
%\newpage
\bibliographystyle{IEEEtran}
\bibliography{IEEEabrv,references}

\end{document}

%% file: body/abstract.tex
Loop closure detection is an essential and challenging problem in simultaneous localization and mapping (SLAM). It is often tackled with light detection and ranging (LiDAR) sensor due to its view-point and illumination invariant properties. Existing works on 3D loop closure detection often leverage the matching of local or global geometrical-only descriptors, but without considering the intensity reading. In this paper we explore the intensity property from LiDAR scan and show that it can be effective for place recognition. Concretely, we propose a novel global descriptor, intensity scan context (ISC), that explores both geometry and intensity characteristics. To improve the efficiency for loop closure detection, an efficient two-stage hierarchical re-identification process is proposed, including a binary-operation based fast geometric relation retrieval and an intensity structure re-identification. Thorough experiments including both local experiment and public datasets test have been conducted to evaluate the performance of the proposed method. Our method achieves higher recall rate and recall precision than existing geometric-only methods.

%% file: body/Introduction.tex
%着重的点 large fov， reverse visit， 
%loop closure fusion 第一段
Loop closure detection, which is also known as place recognition, refers to the capability of identifying a visited place. In the problem of simultaneous localization and mapping (SLAM), the estimated states and trajectories often come with inevitable drift \cite{angeli2008fast}. By identifying the revisited places, a robot can eliminate the drifting error. Moreover, it can also prevent from multiple registration of identical landmarks so that a globally consistent map can be created. 
Vision-based place recognition often suffers from light illumination, weather, or viewing angle.
% A revisited place usually varies from light illumination, weather, or viewing angle, which is difficult to be solved in vision based place recognition. 
Nevertheless, LiDAR is less affected by such environmental changes, hence it has been widely used for place recognition in the recent years. 

\begin{figure}[t]
\begin{center}
\vspace{8pt}
\includegraphics[width=0.99\linewidth]{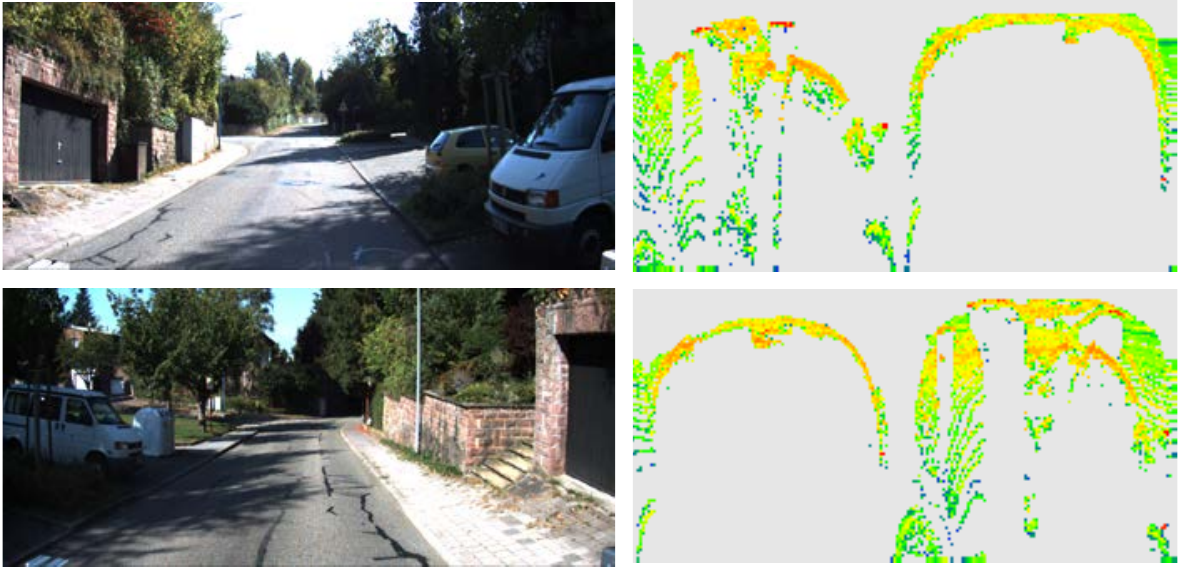}
\captionsetup{justification=justified}
\caption{Example of loop closure detected from KITTI sequence 02. The scenario is challenging due to reverse visit. Our proposed intensity scan context on the right images shows high similarity identifies loop closure.}
\label{fig: title_graph}

\end{center}
\end{figure}

Existing works on place recognition often leverage the matching of 3D descriptors, such as fast point feature histogram (FPFH) \cite{rusu2009fast}, fast laser interest region transform (FLIRT) \cite{tipaldi2010flirt}, and signature of histograms of orientations (SHOT) \cite{tombari2010unique}. These descriptors explore either global or local geometry information such as surface normal or neighbour points distribution, leaving the intensity information unused. A main justification is that intensity information is less straightforward than geometry reading \cite{wang2017non}, since it is affected by not only target surface characteristics (\textit{e.g.}, roughness, surface reflectance), but also acquisition geometry (\textit{e.g.}, distance) and instrument effects (\textit{e.g.}, transmitted energy) \cite{kashani2015review}. However, the intensity channel reveals the reflectance structure of surrounding environment, \textit{e.g.}, retro-reflective material such as metal plate usually returns high value and concrete returns low value. This information is often unique for different places. Moreover, some preliminary works have shown that intensity reading can be effective for place recognition \cite{cop2018delight,guo2019local}.

In this paper, we propose a novel global descriptor, intensity scan context (ISC), that integrates both geometry and intensity characteristics for loop closure detection. We first explain that how intensity information can be distinguishable for a place. Then we propose intensity scan context as a global signature for place recognition. To further improve the efficiency of our algorithm, a two-stage hierarchical place re-identification strategy is proposed, including a binary-operation based fast geometry retrieval and an intensity structure matching. To evaluate the performance of the proposed intensity scan context, our method is tested under different scenarios including outdoor autonomous driving and indoor warehouse robot navigation. The results show that the proposed approach achieves higher recall rate and recall rate than existing geometric-only loop closure detection methods.  

The main contributions of this paper are as follows:
\begin{itemize}
\item We propose a novel global descriptor for 3D LiDAR scan that integrates both geometry and intensity characteristics. 
\item An efficient loop closure detection strategy based on a two-stage hierarchical intensity scan context (ISC) re-identification is proposed. It only costs 1.2 \textit{ms} per query on average.
\item A thorough evaluation on the proposed descriptor, including both local experiment and public datasets test, is conducted.
\end{itemize}

This paper is organized as follows: Section \MakeUppercase{\romannumeral 2} reviews the related works on both vision based and LiDAR based approaches for loop closure detection. Section \MakeUppercase{\romannumeral 3} describes the idea of using ISC for place recognition, followed by the two-stage hierarchical place re-identification. Section \MakeUppercase{\romannumeral 4} shows experimental results and comparison with existing works, followed by the conclusion in Section \MakeUppercase{\romannumeral 5}.

%% file: body/RelatedWork.tex
According to the perception system, existing works on loop closure detection can be categorised as vision-based methods and LiDAR-based methods.
Vision-based approaches are developed for place recognition in the early stage. Those methods often leverage the bag of words model (BoW) that measures the distance of visual words according to a pre-trained visual vocabulary, \textit{e.g.}, FAB-MAP \cite{glover2012openfabmap} and DBoW2 \cite{galvez2012bags}. They are widely used in visual SLAM such as ORB SLAM \cite{mur2017orb} and LDSO \cite{gao2018ldso}. However, image stream is not resistant to light illumination or view-point so that vision-based place recognition is not robust in practice. Although some works aiming to solve the problem of environmental changes have been proposed \cite{anoosheh2019night,naseer2015robust}, they are still limited to some specific scenarios. 

In comparison, due to the high robustness to illumination and view-point changes, LiDAR is subsequently introduced for loop closure detection. Existing works on LiDAR-based place recognition strive for an efficient local descriptor or place signature that can accurately and concisely present a place. One of the most popular local descriptors is the fast point feature histogram (FPFH) \cite{rusu2009fast} which explores the local surface normal of each neighbour point. It is effective in estimating affine transform between two point clouds and is used for place recognition in the later work such as \cite{dube2017segmatch}. Bosse \textit{et al.} propose a probabilistic voting approach based on Gasalt3D descriptor \cite{bosse2013place}. Despite the good performance achieved, the point cloud retrival is inefficient due to the high dimension of the proposed descriptor. 

Re-identification on the local descriptor usually requires key-point extraction and massive local geometry calculation. In comparison, matching on global descriptor is more efficient in place recognition. Rizzini introduces a novel descriptor named GLAROT that encodes the relative geometric position of key-point pairs into a histogram \cite{rizzini2017place}. The experimental results show that it achieves satisfactory recall precision and recall rate. However, building key-point relation is still computationally expensive. Kim \textit{et al.} propose scan context which projects laser scan into global descriptor. The matching of scan context only requires element-wise multiplication so that the query speed is fast. However, the matching precision is not high enough and false positive often occurs during public datasets test.  

The global descriptors are more efficient but the performance is not competitive. However, recent works show that the performance can be improved by integrating intensity characteristics. Guo \textit{et al.} justify that intensity information can be distinctive for places and propose a novel local descriptor called intensity signature of histograms of orientations (ISHOT) that consists of both geometry and intensity information. The place re-identification is solved by a probabilistic voting strategy similar to \cite{bosse2013place}. Despite computationally expensive, the new descriptor outperforms geometrical-only descriptors. This inspires us to propose a more efficient and accurate global descriptor.
% for loop closure detection.

%% file: body/Methodology.tex
In this section, the proposed method is described in detail. We first present the concept of intensity scan and briefly explain that how intensity characteristics can be used for place recognition. Then we present the idea of using intensity scan context as a global 3D descriptor. Furthermore, an efficient two-stage hierarchical intensity scan context based retrieval is introduced, consisting of a fast binary-operation based geometry retrieval and an intensity structure matching.  

\begin{figure}[t]
    \begin{subfigure}{0.99\linewidth}
        \begin{center}
        \includegraphics[width=1.0\linewidth]{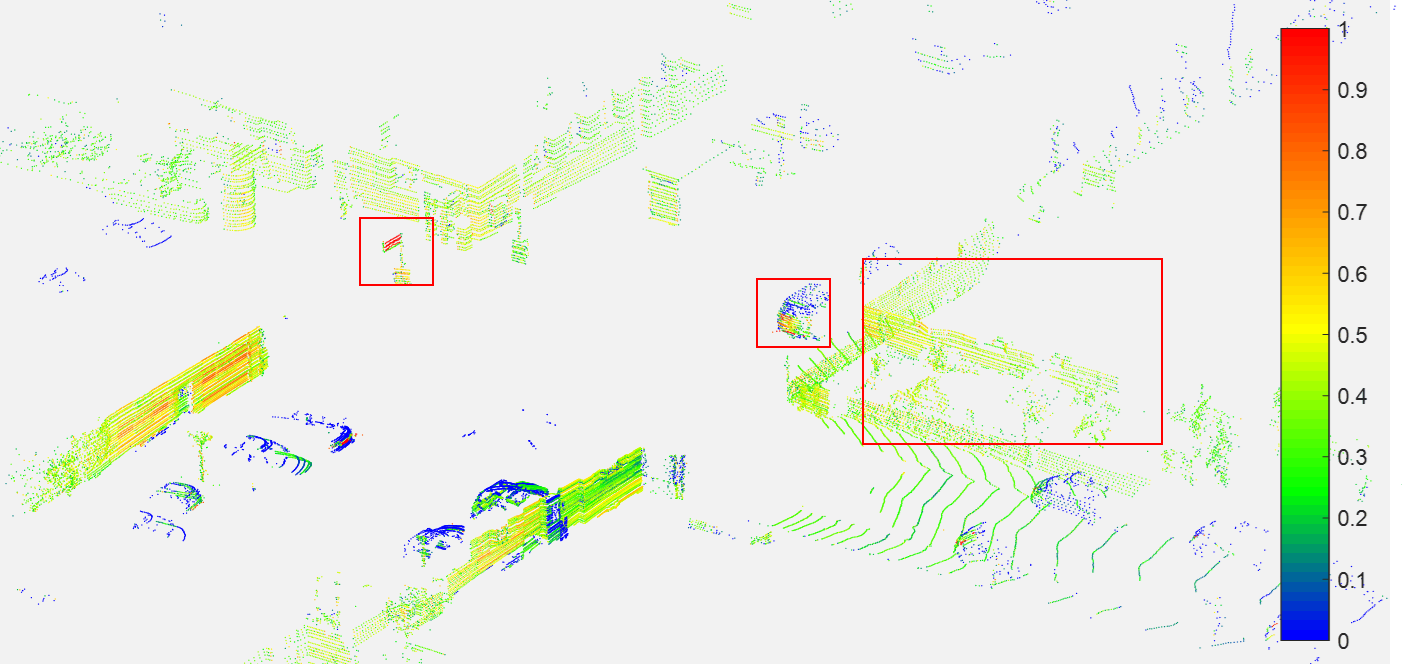}
        \end{center}
        \caption{Intensity scan reading of a crossroad.}
        \label{fig:Front Camrea Image-a}
    \end{subfigure}
    \begin{subfigure}{0.99\linewidth}
        \begin{center}
        \includegraphics[width=1.0\linewidth]{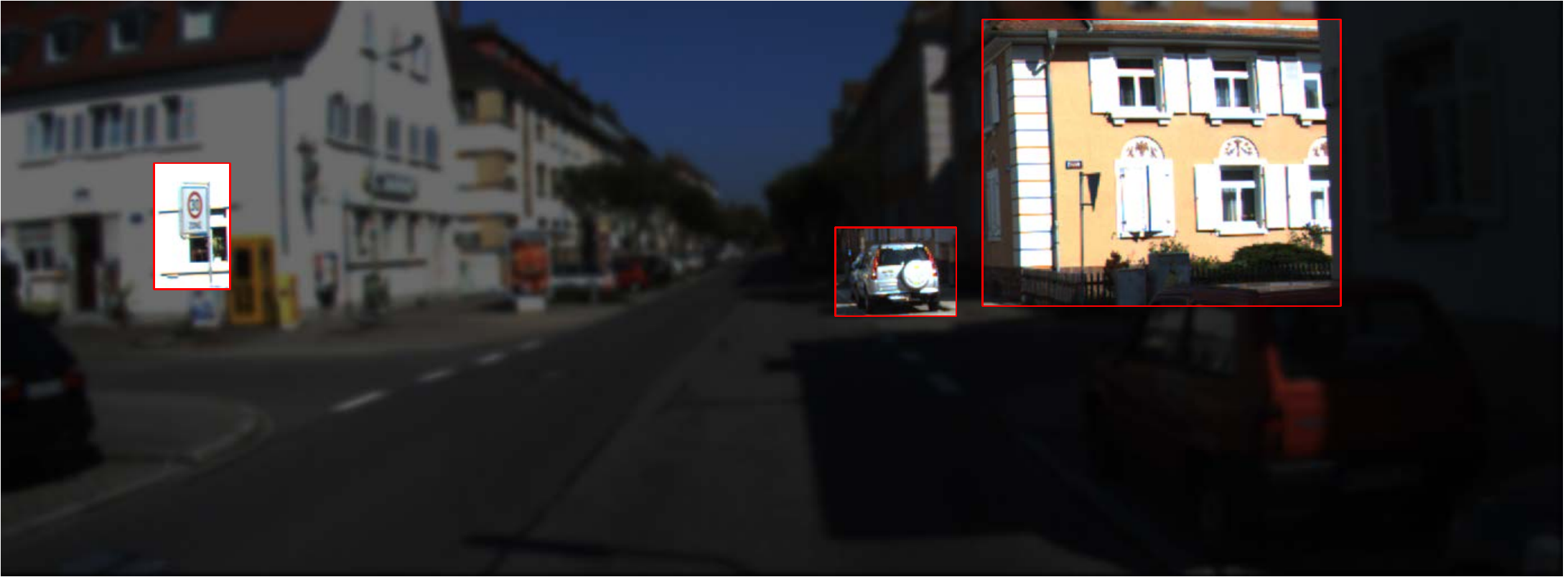}
        \end{center}
        \caption{Camera view of the same place.}
        \label{fig:LiDAR Scan-b}
    \end{subfigure}
\caption{An example of intensity reading from KITTI dataset. The relationship of intensity information and landmarks are highlighted with red rectangles.}
\label{fig:KITTI_intensity_example}
\vspace{-10pt}
\end{figure}

\subsection{Intensity Calibration and Pre-processing}

\begin{figure*}[!t]
    \begin{subfigure}{0.49\linewidth}
        \begin{center}
        \includegraphics[width=0.70\linewidth]{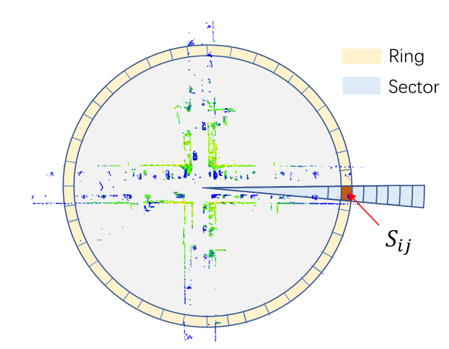}
        \end{center}
        \label{fig:Front Camrea Image-a}
    \end{subfigure}
    \begin{subfigure}{0.49\linewidth}
        \begin{center}
        \includegraphics[width=0.99\linewidth]{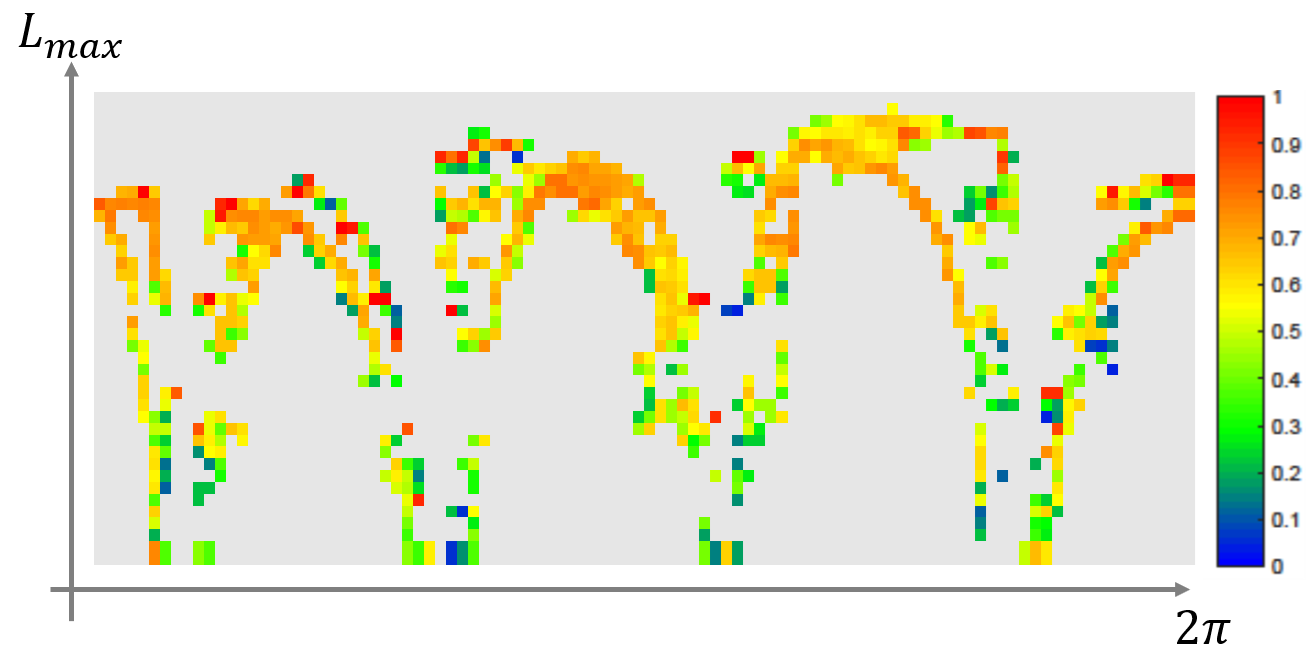}
        \end{center}
        \label{fig:LiDAR Scan-b}
    \end{subfigure}
\caption{A visual illustration of the proposed intensity scan context. Left figure: original point cloud is decomposed into subspace based on geometry characteristics. Right figure: derived intensity scan context by intensity projection on the subspace.}
\label{fig:ISC_result}
\vspace{-10pt}
\end{figure*}
% \begin{figure*}[!t]
% \begin{center}
% \includegraphics[width=0.90\linewidth]{images/ISC_Demo.pdf}
% \end{center}
% \caption{A visual illustration of the proposed intensity scan context. Left figure: original point cloud is decomposed into subspace based on geometry characteristics. Right figure: derived intensity scan context by intensity projection on the subspace.}
% \label{fig:ISC_result}
% \end{figure*}

LiDAR perceives the environment by emitting and receiving laser beam. Generally, the distance value is measured by traveling time while the surface reflectance can be estimated by returned energy level (\textit{i.e.}, intensity). The intensity reading reveals surrounding surface reflectance structure. Existing works on LiDAR have shown that the returned intensity readings vary for different objects \cite{kashani2015review}, \textit{e.g.}, retro-reflective material such as metal plate usually returns high value and concrete returns low value. In Fig. \ref{fig:KITTI_intensity_example}, we show an example from KITTI dataset \cite{geiger2013vision} for demonstration. The point cloud and image present the same place and the intensity structure is interpreted in the first image. We pick 3 landmarks including car, road sign and building respectively and highlight them with red rectangles in both Fig. \ref{fig:KITTI_intensity_example} (a) and Fig. \ref{fig:KITTI_intensity_example} (b). It is observed that the road sign is highly distinguishable with low energy loss while the building structure (concrete) returns medium intensity. Moreover, the reflectance is consistent in object level.

However, the intensity channel is noisy since it is affected by not only target surface characteristics (\textit{e.g.}, roughness, surface reflectance), but also acquisition geometry (\textit{e.g.}, distance) and instrument effects (\textit{e.g.}, transmitted energy) \cite{kashani2015review}. Hence calibration is necessary in order to reduce the disturbance by other factors. Similar to \cite{levinson2014unsupervised,guo2019local}, we calibrate the intensity reading $\eta_r$ with a mapping function $\varphi$:
\begin{equation}
 \eta_{cal}= \varphi(\eta_r,d),
\end{equation}
where $d$ is the distance reading. The mapping function $\varphi$ describes the influence of distance on the received energy and can be collected based on offline experiments. 
Besides remapping, the intensity return from 3D LiDAR is an eight-byte integer, \textit{e.g.}, Velodyne VLP-16. It is re-scaled into [0,1] as a float number for convenience.

In the pratical applications, some pre-processing of the LiDAR scan is necessary in order to remove the redundant information. Based on observation, the LiDAR noise increases with distance. Therefore, the LiDAR reading is firstly filtered by setting a distance threshold $L_{max}$ to remove unreliable points. Moreover, the ground points are often not significant for place recognition so that they are optimized by the method similar to \cite{shan2018lego} in advance. 

\subsection{Intensity Scan Context}

Extraction of local descriptors is often computationally expensive since we need to identify local geometry characteristics such as local norm for every single key-point. Hence to increase the computational efficiency, the intensity information is interpreted as global descriptor in this work. Inspired by scan context \cite{kim2018scan,kim20191} and shape context \cite{belongie2002shape}, we introduce intensity scan context that can efficiently integrate both geometry and intensity characteristics into a global signature. 

Denote the intensity reading $\eta$, geometry reading $[x,y,z]$, and number of points $n$, a LiDAR scan is defined as $\mathcal{P}=\{\mathbf{p}_1,\mathbf{p}_2,\cdots,\mathbf{p}_n\}$ with each point $\mathbf{p_k}=[x_k,y_k,z_k,\eta_k]$ in the local Cartesian coordinate. Each point $p_k$ can be converted into polar coordinate, but only in x-y plane, so that
\begin{equation}
\begin{aligned}
 \mathbf{p_k} &= [\rho_k,\theta_k,z_k,\eta_k],\\
 \rho_k &= \sqrt{{x_k}^2  + {y_k}^2}, \\
 \theta_k &= \arctan \frac{y_k}{x_k}.\\
\end{aligned}
\end{equation}
The point cloud is then segmented by equally diving polar coordinate in azimuthal and radial directions into $N_s$ sectors and $N_r$ rings. Each segment is represented by: 
\begin{equation}
\begin{aligned}
 \mathcal{S}_{ij} = \{\mathbf{p_k}\in \mathcal{P}|\frac{i\cdot L_{max}}{N_r}\leq \rho_k< \frac{(i+1)\cdot L_{max}}{N_r},
 \\
 \frac{j\cdot 2\pi}{N_s}-\pi\leq \theta_k< \frac{(j+1)\cdot 2\pi}{N_s}-\pi\},
 \end{aligned}
\end{equation}
where $i \in [|1,N_s|]$, $j \in [|1,N_r|]$ and the symbol $[|1,N|]$ represents $\{1,2,\cdots,N\}$. The point cloud is then divided into $N_s \times N_r$ subspace. As discussed before, the intensity reading is often consistent for the same object. Since the each subspace is much smaller than whole point cloud, we can assume that the intensity reading does not vary too much. Hence for each subspace, a coding function $\kappa$ can be applied to reduce the intensity dimension. It is defined as:
\begin{equation}
  \begin{aligned}
    \eta_{ij} &= \kappa(\mathcal{S}_{ij}) \\
        &= \max_{p_k\in \mathcal{S}_{ij}} \eta_k.\\
\end{aligned}
\end{equation}
Note that if $\mathcal{S}_{ij}\in \varnothing$ (\textit{i.e.}, no scan data), $\eta_{ij} = 0$. Up to this point, we can generate the intensity scan context $\Omega$ by:
\begin{equation}
  \Omega(i,j) = \eta_{ij}. 
\end{equation}
The global signature $\Omega$ is a 2D matrix that reveals both geometry and intensity distribution of the environment. We pick a LiDAR scan from KITTI dataset \cite{geiger2013vision} for illustration in Fig. \ref{fig:ISC_result}. The left figure is the point cloud from top down view with the sub point cloud space $S_{ij}$ marked in red color. The right figure is the coded intensity scan context with each pixel calculated by intensity coding function.

\subsection{Place Re-identification}
Place recognition targets to match current place $\mathcal{P}_n$ with the previously visited places from the historical database $\mathcal{D} = \{\mathcal{P}_1,\mathcal{P}_2,\cdots,\mathcal{P}_{n-1}\}$. As more places visited, the scale of database $\mathcal{D}$ inevitably increases so that the computational cost grows accordingly. To reduce the computational cost, in this section we propose a two-stage hierarchical intensity scan context retrieval strategy that makes use of fast binary operation to speed up the process of place re-identification. 

\subsubsection{Fast Geometry Re-identification}
Most of 3D descriptors are histogram-based such as unique shape context (USC) \cite{tombari2010unique}, ISHOT \cite{guo2019local}, \textit{etc}. Matching between histograms can be slow in practice since mathematical operation is inevitable, \textit{e.g.}, multiplication of float numbers. In comparison, binary operation (or logical operation) achieves much faster speed than those mathematical operation. Inspired by \cite{cieslewski2016point}, we introduce an efficient binary operation geometry re-identification for fast indexing.  
Given an intensity scan context $\Omega$, its geometry distribution on the local coordinate can be represented as a binary matrix $\mathcal{I}$:
\begin{equation}\label{eq:response}
    \mathcal{I}(x,y) = \left \{
    \begin{aligned}
        false, & ~\text{if}~ \Omega(x,y)=0\\
        true, & ~\text{otherwise}
    \end{aligned} \right.
\end{equation}
For a query intensity scan context $\Omega^q$, a candidate intensity scan context $\Omega^c$ and their binary transform $\mathcal{I}^q$, $\mathcal{I}^c$, the geometry similarity can be derived as:
\begin{equation}
    \varphi_{g}(\mathcal{I}^q,\mathcal{I}^c) = \frac{\textit{XOR}(\mathcal{I}^q,\mathcal{I}^c)}{\left|\mathcal{I}^q\right|} ,
\end{equation}
where $\left|\mathbf{x}\right|$ is the total number of elements in $\mathbf{x}$ and $\textit{XOR}(\mathbf{x},\mathbf{y})$ refers to element-wise exclusive OR operation between matrix $\mathbf{x}$ and $\mathbf{y}$. Because the column vector in intensity scan context represents azimuthal direction, the rotation of laser scan becomes column shift in intensity scan context \cite{kim2018scan}. Hence for place recognition, the viewing angle change can be interpreted as column-shifts of $\Omega$. Therefore, to detect the view-point change, the final score is calculated by:
\begin{equation}
  \Phi_{g}(\mathcal{I}^q,\mathcal{I}^c) = \max_{i\in [|1,N_s|]} \varphi_{g}(\mathcal{I}_i^q,\mathcal{I}^c),
\end{equation}
where $\mathcal{I}_i^q$ is $\mathcal{I}^q$ shifted by $i^{th}$ column. In the meantime, we can identify the best matched $\mathcal{I}_k^q$ with column-shifts of $k$ that can be used to correct viewing angle change. The unmatched pair can be filtered out by setting an empirically determined threshold $\epsilon_g$. In experiments we find that binary matching only costs 0.5 \textit{ms} on a desktop computer that is very computationally efficient. 

\subsubsection{Intensity Structure Matching}
The second stage mainly identifies the intensity similarity between two intensity scan context $\Omega^q$ and $\Omega^c$ by column-wise comparison. Let $\mathbf{v}_i^q$ and  $\mathbf{v}_i^c$ be the $i^{th}$ column of $\Omega^q$ and $\Omega^c$, the score can be found by taking cosine distance:
\begin{equation}
 \begin{aligned}
  \varphi_{i}(\Omega^q,\Omega^c) = \frac{1}{N_s} \sum_{i=0}^{N_s-1}(\frac{\mathbf{v}_i^q \cdot \mathbf{v}_i^c}{\norm{\mathbf{v}_i^q}\cdot \norm{\mathbf{v}_i^c}}).
\end{aligned}
\end{equation}
Similar to geometry matching, we have to correct viewing angle change. Since the viewing angle change $k$ is already identified from geometry re-identification, we compare $\Omega_k^q$ and $\Omega^c$ where $\Omega_k^q$ is $\Omega^q$ shifted by $k^{th}$ column. This also significantly reduces the computational cost because comparison of intensity involves mathematical operation rather than logical operation. The final score is calculated as:
\begin{equation}
 \begin{aligned}
  \Phi_{i}(\Omega^q,\Omega^c) = \varphi_{i}(\Omega_k^q,\Omega^c).
\end{aligned}
\end{equation}
The unmatched pair can be also filtered out by setting an empirically determined threshold $\epsilon_i$. 

\begin{figure}[t]
\begin{center}
\includegraphics[width=0.59\linewidth]{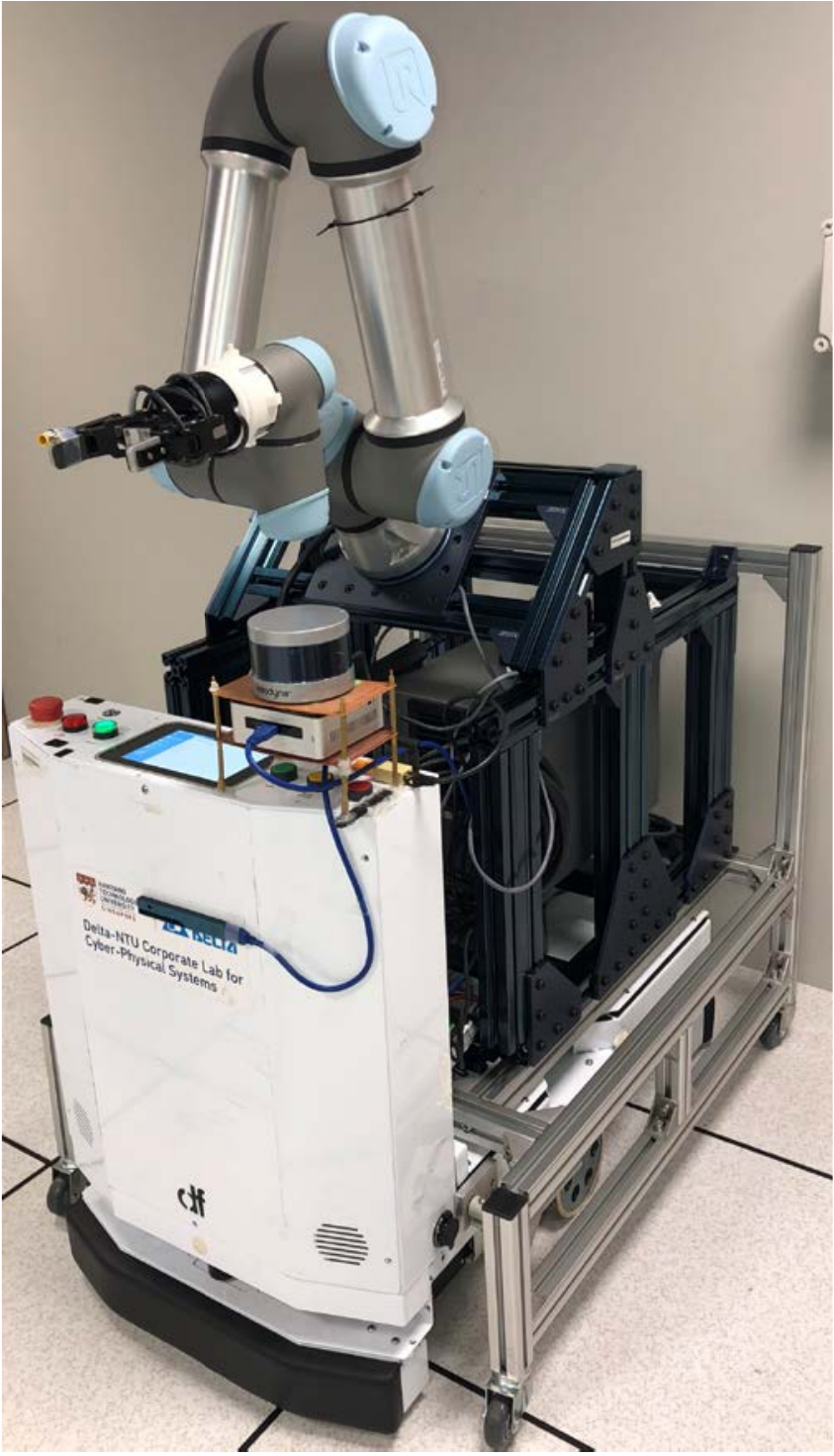}
\captionsetup{justification=justified}
\caption{Autonomous warehouse robot platform used for our experiment.}
\label{fig:warehouse_robot}
\vspace{-10pt}
\end{center}
\end{figure}

\begin{figure*}[!t]
    \begin{subfigure}{0.60\linewidth}
        \begin{center}
        \includegraphics[width=0.85\linewidth]{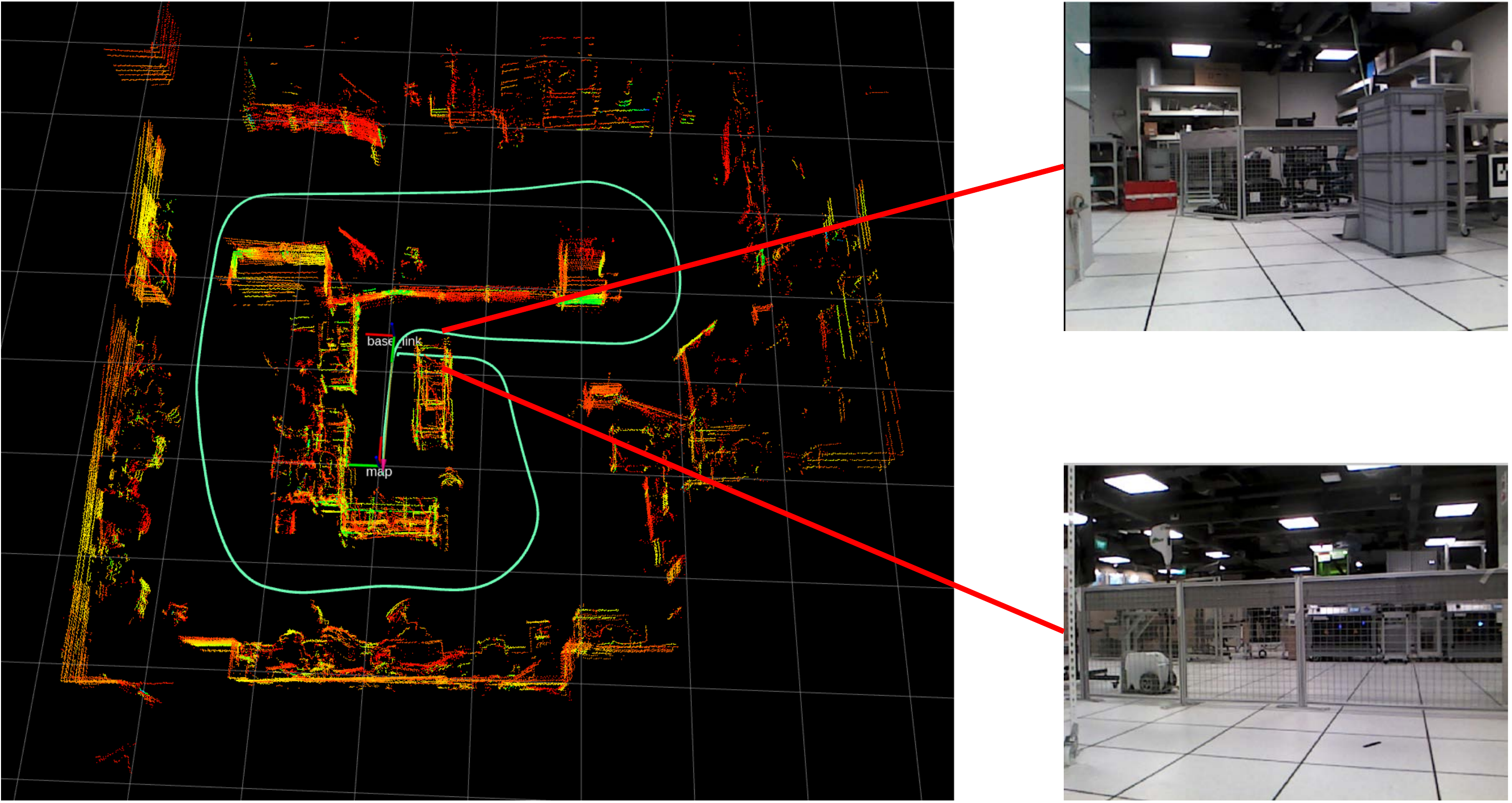}
        \end{center}
        \caption{Loop not detected with visual place recognition.}
        \label{fig:local_experiment_result-a}
    \end{subfigure}
    \hfill
    \begin{subfigure}{0.39\linewidth}
        \begin{center}
        \includegraphics[width=0.8\linewidth]{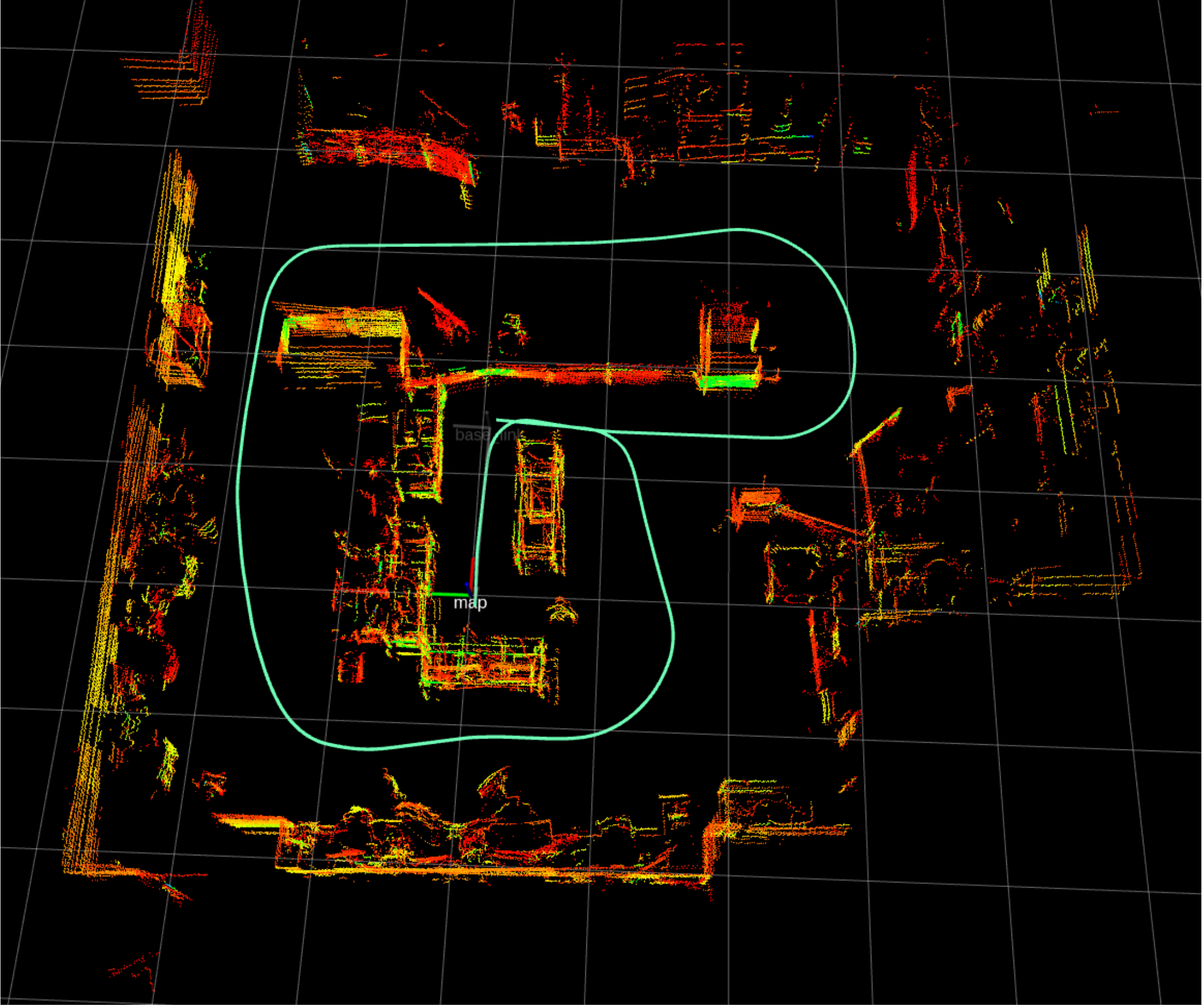}
        \end{center}
        \caption{Loop identified with the proposed approach.}
        \label{fig:local_experiment_result-b}
    \end{subfigure}
\caption{Local experiment result of the proposed algorithm.}
\label{fig:local_experiment_result}
\vspace{-10pt}
\end{figure*}

\subsection{Consistency Verification}
As discussed in previous section, global descriptor is a highly simplified representation of original point cloud. Hence it is inevitable to have some features ignored, which can lead to false positive. Therefore it is necessary to check consistency before closing the loop.
\subsubsection{Temporal consistency check} 
In a SLAM system, it is observed that the occurrence of single loop closure often implies high similarity on the neighbour LiDAR scans since the sensor feedback is continuous in time \cite{milford2012seqslam}. We can verify the loop closure by measuring the temporal consistency:
\begin{equation}
    P(\mathcal{P}_m,\mathcal{P}_n) = \frac{1}{N} \sum_{k=1}^{N} (\Phi_{g}(\mathcal{I}_{m-k},\mathcal{I}_{n-k})+ \Phi_{i}(\Omega_{m-k},\Omega_{n-k})),
    \label{eqn: overall likelihood}
\end{equation}
where $N$ is the number of frames included for temporal consistency verification. Note that in case of reverse visit (viewing angle change detected in geometry re-identification), $\mathcal{I}_{m-i}$ becomes $\mathcal{I}_{m+i}$ accordingly. The loop candidate can be accepted by taking threshold $\xi$ on the final temporal consistency score.

\subsubsection{Geometrical consistency} The geometrical consistency verifies the raw scan-to-scan similarity. Similar to \cite{guo2019local}, FPFH features are extracted and matched to get an initial guess of the rigid transform matrix. Starting from this initial estimate, iterative closest point (ICP) \cite{besl1992method} is applied to find minimal distance error between the query scan and candidate scan.

%% file: body/Experiments.tex
\subsection{Experiment Setup}

In this section we present preliminary results from our proposed method, including both indoor warehouse robot navigation and outdoor autonomous driving experiments. The proposed method is implemented in C++ and is integrated to robot operating system (ROS) on an Intel NUC mini computer. For indoor warehouse navigation, an autonomous guided vehicle (AGV) equipped with Velodyne VLP-16 and Intel Realsense r200 is used for the experiment. As shown in Fig. \ref{fig:warehouse_robot}, the robot is developed for warehouse manipulation tasks such as packaging and transportation which require high accuracy on long-term localization. The maximum speed of AGV is 1 \textit{m/s}. For outdoor autonomous driving, KITTI dataset \cite{geiger2013vision} is used for evaluation. The proposed place recognition approach is used to reduce localization drifts and improve the mapping accuracy. A list of parameters used in the experiment is shown in Table \ref{fig:parameter_list}.

\begin{table}[!b]
\begin{center}
\begin{tabular}{ccc}
\toprule

Parameter  & Description&  Value \\ 
\midrule
$L_{max}$   &  Maximum radius        &   50         \\ 
$N_s$   &  Number of sectors        &      20        \\ 
$N_r$     &  Number of rings        &      60     \\  
$\epsilon_g$     &  Geometry matching threshold        &      0.9     \\  
$\epsilon_i$     &  Intensity matching threshold        &      0.92     \\  
$N$     &  Frames used in temporal consistency check        &      5     \\  
$\xi$     &  Temporal consistency checking threshold        &      1.8     \\  
\bottomrule
\end{tabular}
\caption{Parameter List.}
\label{fig:parameter_list}
\end{center}
\end{table}

\subsection{Experiment on Autonomous Robot}
The robot is tested in real warehouse environment consisting of operating machines, shelves, human, \textit{etc}. LiDAR odometer is estimated based on feature points matching via point cloud library (PCL) \cite{Rusu_ICRA2011_PCL} and the robot trajectory (front-end SLAM) are collected from the fusion of both wheel odometer and LiDAR odometer. To illustrate, we simulate a common task where the robot leaves the docking station to fetch materials and come back in a reverse direction. Our approach is compared with vision-based approach using a front-mounted camera. In particular, we use DBoW2 \cite{galvez2012bags} which is implemented in ORB SLAM \cite{mur2017orb}. The result is shown in Fig. \ref{fig:local_experiment_result} with the estimated trajectory plotted in green. For DBoW2, the reverse visit changes the view angle significantly so that there is not enough similarity to conclude loop closure. Hence the trajectory collides with shelves according to the final result, which is incorrect. In comparison, our proposed method is able to identify the revisited place easily. The estimated trajectory and the mapping are much more reasonable. This is due to the view angle invariant property of our proposed two-stage hierarchical ISC-based retrieval.

\subsection{Evaluation on Public Dataset}
To further demonstrate the performance of the proposed method, we test the algorithm on the KITTI dataset which is commonly used for place recognition. KITTI dataset is collected from an autonomous car equipped with various perception systems including front-mounted cameras, Velodyne HDL-64E LiDAR, GPS, \textit{etc}. The recorded scenarios are challenging for loop closure detection due to similar architectures and dynamic environment. The proposed method is tested with multiple recordings such as sequence 00, 02 and 05. Sequence 00 and 05 are commonly used for place recognition since most of loop closure places are forward visited. Sequence 02 contains both forward and reverse visit and is considered as a more challenging scenario.

\begin{table}[!b]
\begin{center}
\renewcommand{\arraystretch}{1.4}
\begin{tabular}{c|c|c|c}
\hline 
Dataset  & Approaches & Precision (\%) &  Recall Rate (\%) \\ \hline 
\multirow{3}{*}{sequence 00} & Kim \cite{kim2018scan}              & 100    &   87   \\ \cline{2-4}
                             & GLAROT3D \cite{rizzini2017place}              & 86    &   40  \\ \cline{2-4}
                          & Cieslewski \cite{cieslewski2016point}              & 92    &   80  \\ \cline{2-4}
                          & G{\'a}lvez-L{\'o}pez \cite{galvez2012bags}         & 100    &   92   \\ \cline{2-4}    
                          & \textbf{\textit{Proposed}}             & \textbf{100}    &   \textbf{90.2}   \\ 
                          \hline 
\multirow{3}{*}{sequence 02} & Kim \cite{kim2018scan}              & 90    &   73   \\ \cline{2-4}
                          & G{\'a}lvez-L{\'o}pez \cite{galvez2012bags}         & 100    &   80.6   \\ \cline{2-4}    
                          & \textbf{\textit{Proposed}}              & \textbf{98}    &   \textbf{91}   \\ 
                          \hline 
\multirow{3}{*}{sequence 05} & Kim \cite{kim2018scan}              & 100    &   90   \\ \cline{2-4}
                            & Cieslewski \cite{cieslewski2016point} & 93    &   60   \\ \cline{2-4}
                            & GLAROT3D \cite{rizzini2017place}              & 80    &   80  \\ \cline{2-4}
                          & G{\'a}lvez-L{\'o}pez \cite{galvez2012bags}         & 100    &   87.6   \\ \cline{2-4}    
                          & \textbf{\textit{Proposed}}              & \textbf{100}    &   \textbf{91.2}   \\ 
                          \hline 
\end{tabular}
\caption{Comparison with existing methods.}
\label{table:comparison.}
\end{center}
\end{table}

\begin{figure*}[!t]
\begin{center}
\includegraphics[width=0.95\linewidth]{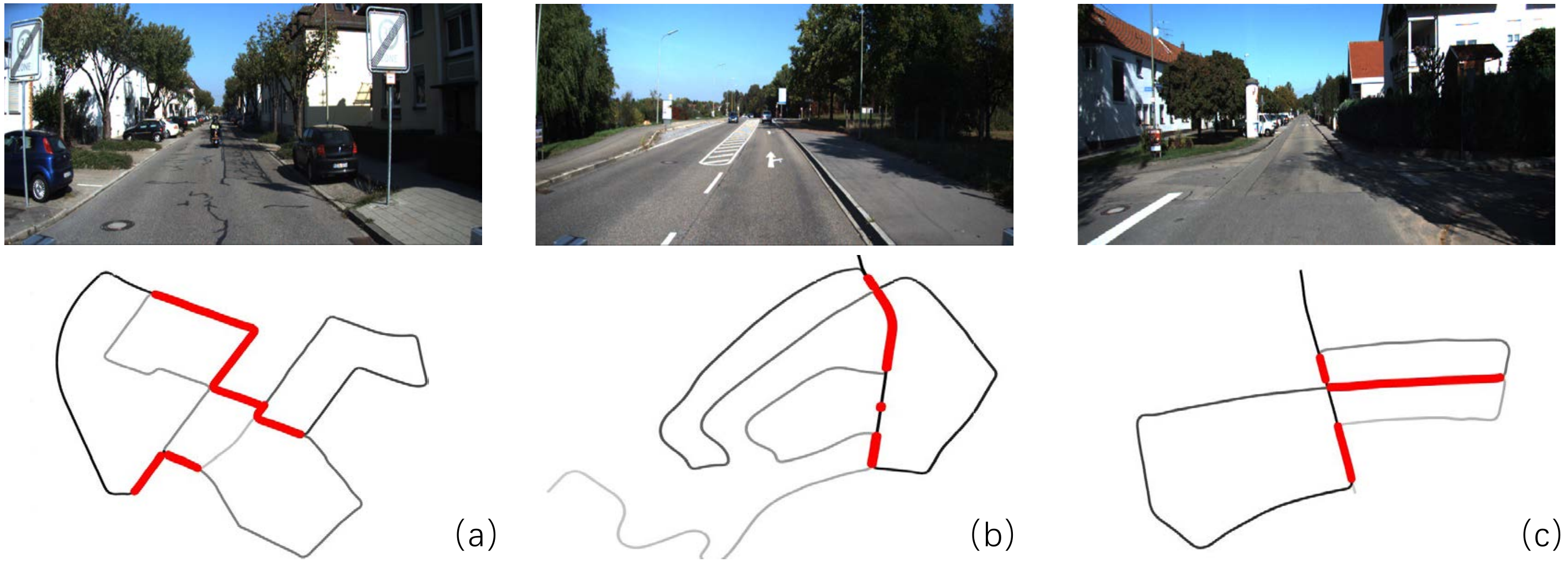}
\captionsetup{justification=justified}
\caption{Loop closure detection result. (a) KITTI sequence 00. (b) KITTI sequence 02. (c) KITTI sequence 05.}
\label{fig: precision_reall}
\vspace{-10pt}
\end{center}
\end{figure*}

Recall rate and precision are collected from each test. Precision is the percentage of successful pairing and recall rate is the reported loop pairs against total loop pairs. Higher recall rate can significantly reduce drifting error and higher recall precision can prevent wrong registration of points into map. The number of total loop closure is collected based on GPS since the test-field is large in scale. The results of our proposed approach can be found in Fig. \ref{fig: precision_reall}, with the GPS trajectory plotted from light to dark with time going. Each of the loop closure place detected is marked in red.

The results are also compared with existing works such as \cite{kim2018scan,rizzini2017place,cieslewski2016point} that are commonly used in SLAM applications. We also include the state-of-the-art vision-based place recognition methods such as \cite{galvez2012bags} for comparison. The experimental results of \cite{kim2018scan,rizzini2017place,cieslewski2016point} are collected from the respective paper due to unavailability of source code and \cite{galvez2012bags} is tested with local experiment. The results are listed in Table \ref{table:comparison.}. Compare to vision-based approach, our approach achieves competitive precision and recall rate on both sequence 00 and sequence 05. On more challenging dataset sequence 02, our approach achieves much higher recall rate. This is because in sequence 02 vision based approach fails to identify the reverse visit so that the recall rate drops significantly. However, false positive is reported for LiDAR based approach such as our method and \cite{kim2018scan}. The failure case comes from non-residential area where both sides of roads are trees so that the geometry and intensity characteristics are very limited for place recognition. Compared to LiDAR based approach such as \cite{kim2018scan,rizzini2017place,cieslewski2016point}, we achieve both higher recall precision and recall rate across all three datasets. Moreover, the proposed intensity scan context only costs 1.2 \textit{ms}/query which is very efficient in practice.

%% file: body/Conclusion.tex
In this paper, we present a robust loop closure detection approach by integrating both geometry and intensity information. Existing works on LiDAR-based loop closure detection mainly leverage the geometric-only descriptor and ignore intensity reading. Inspired by the recent researches on LiDAR intensity, we argue that the intensity information can be effective for place recognition and propose a global 3D descriptor named intensity scan context. To reduce the computational cost, an efficient two-stage hierarchical intensity scan context retrieval is proposed, consisting of a fast binary-operation based geometry indexing and an intensity structure re-identification. It costs only 1.2 \textit{ms} per query on a normal computer in practice. Thorough experiments have been conducted including local run test with autonomous warehouse robot and public dataset test to evaluate our proposed method. The results show that our proposed method achieves competitive recall precision and recall rate compared to the state-of-the-art methods.